# Uncertainty Quantification Framework for Aerial and UAV Photogrammetry through Error Propagation

Debao Huang[1,2,3] and Rongjun Qin[1,2,3,4,*]

[1] Geospatial Data Analytics Laboratory, The Ohio State University, Columbus, USA
[2] Department of Civil, Environmental and Geodetic Engineering, The Ohio State University, Columbus, USA
[3] Department of Electrical and Computer Engineering, The Ohio State University, Columbus, USA
[4] Translational Data Analytics Institute, The Ohio State University, Columbus, USA

**Abstract** — Uncertainty quantification of the photogrammetry process is essential for providing per-point accuracy credentials of the point clouds. Unlike airborne LiDAR, whose accuracy generally remains consistent with objects with varying geometric complexity, the accuracy of photogrammetric point clouds is rather object/scene-dependent, as it relies on algorithm-derived measurements. Generally, errors of the photogrammetric point clouds propagate through a two-step process: Structure-from-Motion (SfM) with Bundle adjustment (BA), followed by Multi-view Stereo (MVS). While uncertainty estimation in the SfM stage has been well studied using the first-order statistics of the reprojection error function, that in the MVS stage remains largely unsolved and non-standardized, primarily due to its non-differentiable and multi-modal nature (i.e., from pixel values to geometry). In this paper, we present an uncertainty quantification framework closing this gap by associating an error covariance matrix per point accounting for this two-step photogrammetry process. Specifically, to estimate the uncertainty in the MVS stage, we propose a novel, self-calibrating method by taking reliable n-view points ($n \geq 6$) per-view to regress the disparity uncertainty using highly relevant cues (such as matching cost values) from the MVS stage. Compared to existing approaches, our method uses self-contained, reliable 3D points extracted directly from the MVS process, with the benefit of being self-supervised and naturally adhering to error propagation path of the photogrammetry process, thereby providing a robust and certifiable uncertainty quantification across diverse scenes. We evaluate the framework using a variety of publicly available airborne and UAV imagery datasets. Results demonstrate that our method outperforms existing approaches by achieving high bounding rates without overestimating uncertainty. More information, including test cases and sample datasets, can be found at: https://github.com/GDAOSU/UncertaintyQuantification.

**Index Terms**—Uncertainty quantification, error propagation, multi-view stereo, photogrammetry, point cloud, unmanned aerial vehicle.

## 1. INTRODUCTION

Aerial and unmanned aerial vehicle (UAV) photogrammetry are widely used to generate high-resolution 3D data for various applications [1-5], offering advantages such as low cost, reduced labor, and flexible data acquisition. Quantifying the uncertainty of these 3D datasets is increasingly important to assume their trustworthy use in practice. Currently, most applications rely on check points to assess the accuracy of 3D models [6-8], which provides only sparse and discrete validation across the entire 3D product. While this approach is generally sufficient for typical mapping projects, it does not provide per-point accuracy estimates required for more sophisticated use cases, such as in simulation, synthetic environments [9, 10], and modeling of mission-critical infrastructures, where error modeling is critical. Therefore, if the uncertainty of photogrammetric point clouds can be accurately and reliably estimated, it can support broader adoption in downstream tasks.

Aerial and UAV photogrammetry [11, 12] is typically accomplished by a two-step process: Structure-from-Motion (SfM) and Multi-view Stereo (MVS). SfM can be regarded as the automated form of Aerotriangulation [13, 14], while requiring less controlled setups such as prior camera calibration and extensive ground control. Over the past decades, SfM has evolved from geometry-based methods [15, 16] to data-driven, deep learning (DL)-based approaches [17, 18]. Despite these advancements, incremental SfM remains dominant in current practices of aerial and UAV photogrammetry [11, 12] due to its capability to handle large-scale, high-resolution datasets [19] and robustness to outliers. Incremental SfM begins with an initial two-view reconstruction using a carefully selected image pair [20], which is then iteratively extended by adding new images [21, 22] and triangulating new 3D points. Bundle adjustment (BA) [23] is applied throughout the incremental reconstruction to jointly refine 3D points, camera poses, and calibration parameters. The resulting sparse reconstruction can be georeferenced if Global Navigation Satellite System (GNSS), Inertial

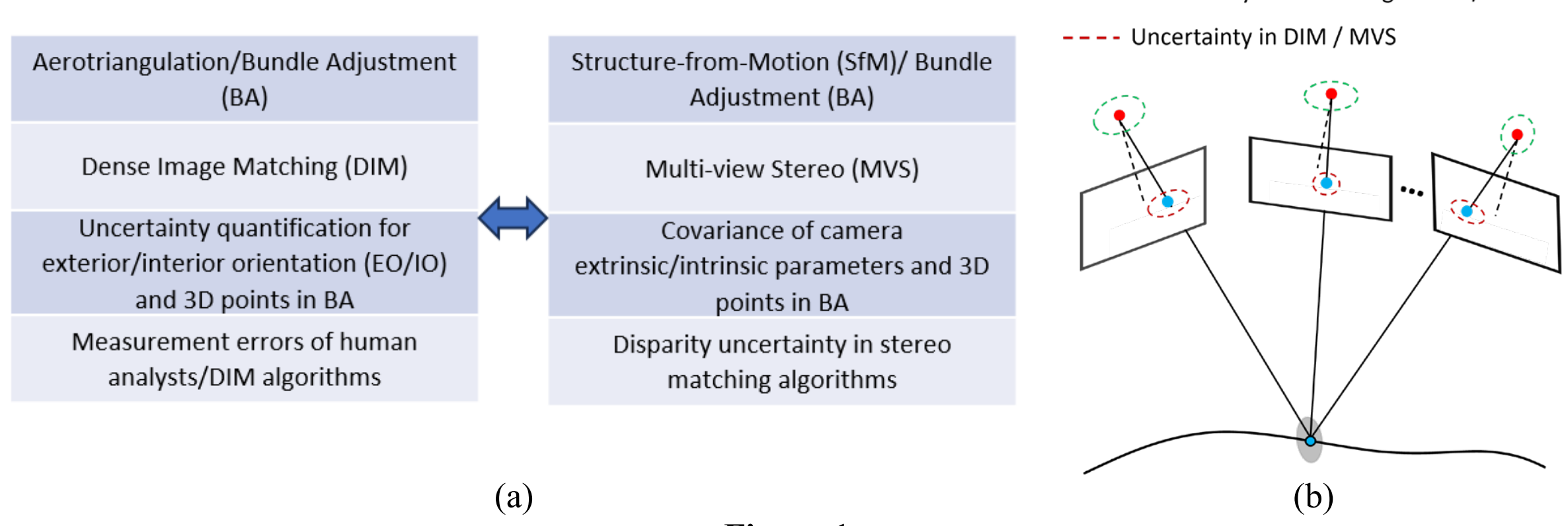

**Figure 1**

(a) Common terminology and task conventions. (b) Overview of error propagation process in uncertainty quantification framework. We propose a novel method for estimating the uncertainty in MVS stage, which can be propagated to error covariance matrix of 3D points.

Measurement Unit (IMU) data or ground control points (GCPs) are available. In the second stage, MVS takes the SfM output and extends stereo matching (of two views) across multiple views to generate dense 3D point clouds or digital surface models (DSM), a process also known as Dense Image Matching (DIM) [24-26]. We list the common terminology conventions and research tasks in **Figure 1 (a)** and consider SfM/Aerotriangulation and MVS/DIM largely interchangeable in this work. Most of the current practices employ depth-map fusion methods due to their ability to provide finer geometry and better scalability [27-29]. These methods first perform stereo matching between each image and its neighboring views to generate pairwise disparity/depth maps, which are subsequently fused to generate a dense 3D point cloud for each image. All points clouds can either be simply merged or further fused to form a unified 3D reconstruction. Compared to ground-based photogrammetry, aerial and UAV photogrammetry often achieve higher reconstruction accuracy due to several factors. First, photogrammetric blocks with designed overlaps ensure high multi-view co-occurrence and sufficient parallax for triangulation, providing more favorable geometric conditions for SfM [30, 31]. Second, the subsequent MVS stage produces abundant multi-view points, of which the redundant measurements mitigate noises.

Nevertheless, this two-step process is sensitive to different scene configurations, such as illumination variations, low-texture areas, or repetitive patterns, which pose significant challenges for uncertainty quantification of the resulting 3D point clouds. Uncertainty in the SfM stage has been well studied in the photogrammetric framework [13, 32-35] through BA, where, supported by the Gauss-Markov theorem [36, 37], uncertainty of the camera poses and calibration parameters can be estimated using the first-order derivative of the reprojection error function under a local linear approximation. On the other hand, the estimation of uncertainty in the MVS stage, which encompasses the measurement uncertainty in the pairwise stereo matching process (i.e., disparity uncertainty), remains unresolved and non-standardized due to its inherent complexity: first, the measurement process is performed through various algorithm-based pixel correlation techniques. Second, MVS algorithms involve a search procedure to identify corresponding pixel values and convert them into disparity geometry. This search procedure is essentially a multi-modal process that is non-differentiable. Third, the inherent variability of pixel values affected by environmental factors such as lighting, reflections, and shading, etc., further complicates the stereo measurement process. While many studies have investigated uncertainty in photogrammetry across various aspects, such as self-calibrating BA [32, 38, 39] and manual point measurement [40, 41], the uncertainty in automated stereo matching algorithms has not yet been sufficiently characterized, as the modern measurements are mostly generated by algorithms rather than human analysts. Recent studies focus on either disparity confidence estimation [42-44], which provides only a 0 – 1 scalar not suitable for propagation, or on pairwise disparity uncertainty estimation performed independently [45-47], which nevertheless neglects the multi-view information available within MVS. Consequently, a fully automated, algorithm-based scheme capable of producing per-point covariance matrices for any point of interest in dense 3D point clouds is critical for modern photogrammetry, as such metadata can substantially enhance the usability and reliability of photogrammetric point clouds.

In this paper, we present an uncertainty quantification framework tailored to aerial and UAV photogrammetry, which computes a 3×3 error covariance matrix for each 3D point that follows the error propagation path of the two-step photogrammetry process, as illustrated in **Figure 1 (b)**. Our framework is capable of predicting both absolute and relative uncertainty. Absolute uncertainty characterizes the global misalignment of the entire point cloud with respect to its true geographic position, whereas relative uncertainty varies from point to point, mainly due to errors in pixel correspondence within matching algorithms. While we estimate relative uncertainty in this work, the framework can also estimate absolute uncertainty when GNSS/GCP information is available. To this end, we propose a novel method to estimate uncertainty in the MVS stage based on our preliminary work [48]. Specifically, we utilize the matching cost values as an indicator that is highly correlated with measurement quality, offering the advantages of being pixel-wise with a similar resolution to the 3D point clouds, while reducing the computational cost than using the full cost volume. We use this cue to regress the disparity uncertainty using the n-view points ($n \geqslant 6$) from the MVS stage. These points have been validated in our previous study [48] to be stable, accurate, and sufficient in aerial and UAV photogrammetry, making them suitable as pseudo-check points. The self-contained n-view points ($n \geqslant 6$) are used to self-calibrate the magnitude of disparity uncertainty for each stereo pair, which cannot be reflected by the pairwise cue alone. Our approach is thus self-supervised and explores the full potential of the multi-view information. The main contributions of this paper are twofold:

1) We propose a novel method for estimating uncertainty in the MVS stage, which uses pairwise matching cost cues and reliable n-view points ($n \geqslant 6$) within MVS to regress disparity uncertainty for each stereo pair. Our method does not require external supervision and achieves better generalization than existing supervised approaches.
2) We present an uncertainty quantification framework that provides a statistically grounded propagation of uncertainty from SfM and MVS to the error covariances of 3D point clouds. We conduct a comprehensive evaluation of the framework on publicly available airborne and UAV datasets, assessing both disparity uncertainty and uncertainty of 3D point clouds. To the best of our knowledge, this is the first in-depth assessment of its kind.

The remainder of this paper is organized as follows: **Section 2** provides an overview of related works; **Section 3** details our method for estimating the uncertainty in the MVS stage and presents the uncertainty quantification framework; **Section 4** describes the dataset preparation, experimental setup, evaluation results, and sensitivity analysis; **Section 5** concludes the paper and outlines future work directions.

## 2. RELATED WORK

This section provides an overview of existing methods for uncertainty estimation in the individual stages of SfM and MVS respectively in **Section 2.1** and **Section 2.2**, with a particular focus on MVS. We then review recent efforts on unified uncertainty quantification frameworks that adhere to the error propagation path of the two-step photogrammetry process in **Section 2.3**.

### 2.1. Uncertainty Estimation of Poses in SfM/Photogrammetry

Accuracy is a combination of precision (repeatability of measurements) and trueness (closeness of their average to true value). While internal precision metrics (i.e., residuals from BA) [39] can be used to evaluate the internal reliability of the system, the accuracy of photogrammetric point clouds can be assessed using independently surveyed check points [49-51] when available. For example, Statistics such as Linear Error at 90% (LE90) and Circular Error at 90% (CE90) [52, 53] can be derived for accuracy assessment. Both CE90 and LE90 represent the radius within which 90% of the positional errors fall, with LE90 measuring error along a single axis and CE90 representing the combined horizontal positional error as a circle. Besides sampled check points, geo-referenced LiDAR-derived point clouds can also be used to serve as reference data for accuracy assessment of photogrammetric point clouds in many benchmarks [54, 55], providing more comprehensive per-point accuracy evaluations.

While accuracy measures the closeness to ground truth, uncertainty reveals the reliability of results by accounting for error propagation in the photogrammetry process. It has been well practiced in many mapping applications with GCPs [56, 57] to generate georeferenced precision maps that inform absolute positional uncertainty. Without GCPs, the quality of SfM reconstruction can be evaluated by best propagating the uncertainties of the input measurements (image observations) to the estimated parameters (3D points, camera poses, and calibration parameters) through the projection function [22]. In practice, the nonlinear projection function is linearly approximated using first-order statistics derived from the Jacobian matrix, from which the information matrix is computed. The covariance matrices of the estimated parameters are then obtained as the inverse of the information matrix (i.e., approximated Hessian Matrix) [58, 59]. Since SfM reconstructions are generally determined only up to an unknown similarity transformation (i.e., gauge ambiguity), early gauge-free methods [60-62] employed the Moore-Penrose (M-P) pseudoinverse [63] of the information matrix to resolve this gauge freedom. However, M-P inversion is computationally expensive, especially for large-scale reconstructions. Subsequent works [62, 64] proposed the use of Taylor Expansion (TE) inversion as a replacement for M-P inversion, significantly reducing computational cost. More recently, a method called NBUP [65] is proposed to eliminate any approximation by combining nullspace computation SfM with the constrained Gauss-Markov model, which achieves both high accuracy and computational efficiency for uncertainty estimation in large-scale SfM.

### 2.2. Uncertainty Estimation in Stereo/MVS

MVS extends pairwise stereo matching to multiple images to reconstruct dense 3D point clouds [66] and DSMs [24, 25, 67], among which the multi-ray stereo matching method has been widely applied in aerial and UAV photogrammetry [68]. Uncertainty estimation in such MVS methods aggregates disparity/depth uncertainties across multiple stereo pairs, i.e., the measurement uncertainties in the pairwise stereo matching process. While confidence estimation in stereo matching has been more extensively studied through both handcrafted [42, 43, 69-72] and learning-based approaches [44, 73-82], it is only a probabilistic measure of how "good" a predicted measurement is, which does not indicate the range over which a predicted measurement can be considered "good". On the other hand, the explicit quantification of uncertainty has received comparatively less attention. A series of studies [47, 83] introduces the Total Variance (TV) metric, which measures local disparity oscillation and classifies pixels into discrete TV classes. Each TV class is then mapped to a corresponding disparity uncertainty value, which is learned from GT disparity maps via an expectation-maximization (EM) process. However, the TV-based approach suffers from limitations such as discretized uncertainty predictions and poor generalization to out-of-domain datasets. Recent Deep learning (DL)-based approaches, such as SEDNet [46], jointly estimate disparity and uncertainty for each stereo pair. SEDNet employs a Convolutional Neural Network (CNN) for disparity prediction and a dedicated Multilayer Perceptron (MLP) to regress disparity uncertainty. The loss function adds a Kullback–Leibler (KL) divergence term to align the predicted uncertainty distribution with the actual error distribution. Tonchev, et al. [84] apply a strategy similar to SEDNet for uncertainty estimation, but replace the stereo matching model with MobileStereoNet[85]. Their uncertainty prediction is essentially the product of a confidence measure and a fixed factor representing the maximum expected error. DUE-MVSNet [86] inputs differences between multi-stage depth maps into an MLP network to regress the uncertainty map. Another type of DL-based approaches [45, 87, 88] leverage Bayesian Neural Networks (BNNs) to predict both depth and uncertainty by modeling the posterior distribution of network parameters, approximated via variational inference by minimizing KL divergence. Depth uncertainty is then inferred by marginalizing over the learned parameter distribution. Other approach [89] employs a deep evidential regression framework to estimate the evidential distribution parameters for the disparity uncertainty. Although DL-based methods show promising results in uncertainty prediction, their uncertainty estimation modules are tightly coupled with specific deep stereo matching architectures, limiting their adaptability to other stereo matching algorithms. Their computational demands – particularly GPU memory requirements – limit these models to being trained and evaluated on synthetic, indoor, or ground-level datasets with small baselines, while their performance on high-resolution airborne and UAV datasets with large baselines remains largely untested.

Importantly, most of these methods are designed for pairwise stereo matching alone and do not fully leverage the multi-view information inherent in MVS. This limitation is especially relevant in airborne and UAV imagery with sufficient multi-view overlap. In contrast, our proposed method gives uncertainty bounds in real metrics for photogrammetric measurement. It is self-supervised and explicitly incorporates the multi-view information in MVS by exploiting n-view points ($n \geqslant 6$) for robust disparity uncertainty regression, which is a substantial departure from existing methods.

### 2.3. Uncertainty Quantification for Photogrammetric 3D Reconstruction

Numerous studies have laid the foundation for uncertainty estimation in photogrammetry. Early works investigate systematic image errors caused by sensor and film deformation and their effects on self-calibrating BA. Other studies focus on how network geometry influences the accuracy potential of final results, including the density and distribution of control points [32, 33, 39]. Several works also expand the investigation to the measurement process itself, including the analysis of measurement errors introduced by image matching performed by human analysts [40, 41]. With the development of computer-based techniques, studies have emerged on the uncertainty in automatic aerotriangulation pipelines, including the uncertainty of exterior orientation parameters and their effects on establishing correspondences of tie points across multiple images [13]. While uncertainty estimation for SfM has been well studied in the context of photogrammetric aerotriangulation, the uncertainty of MVS – representing measurement errors in automated stereo matching algorithms – has not yet been characterized, as measurements in modern workflows are largely generated by automated algorithms rather than human analysts. Comparatively few recent studies have focused on quantifying the uncertainty of final 3D point clouds by propagating uncertainties throughout this two-step process. Rodarmel, et al. [90] present the first framework to propagate uncertainties from SfM and MVS to the error covariances of 3D point clouds derived from UAV imagery. Specifically, it adapts the TV-based method [47, 83] to estimate disparity uncertainties. The work was evaluated using proprietary UAV datasets; its performance on publicly available airborne and UAV datasets remains unverified. Furthermore, the study focuses solely on the uncertainty of 3D point clouds without providing a quantitative assessment of disparity uncertainty. Nocerino, et al. [91] present an uncertainty quantification framework for evaluating 3D models of coral reefs in underwater photogrammetry. However, it propagates only the uncertainty from SfM and does not explicitly model and propagate uncertainty in MVS, instead using the number of stereo pairs merely as a damping parameter to adjust final uncertainty estimates. For satellite imagery, Mundy and Theiss [92] propose an uncertainty quantification framework for DSMs. Rather than explicitly propagating uncertainty from MVS, the method assigns probabilistic weights to the 3D points of each stereo pair based on consistency between two stereo matching runs with reversed image order. The final error covariances are derived through a weighted least-squares algorithm during the fusion of 3D points into DSM grids.

Overall, most existing uncertainty quantification frameworks lack an appropriate approach in the MVS stage. In this paper, we present an uncertainty quantification framework that follows the error propagation path of the two-step photogrammetry process. The most relevant work to ours is that of Rodarmel, et al. [90], which adapts the TV-based method in their framework. Compared to their framework, we integrate our proposed method for uncertainty estimation in the MVS stage, benefiting from the use of multi-view information. We demonstrate this improvement through evaluations of both disparity uncertainty and the uncertainty of 3D point clouds using publicly available airborne and UAV datasets.

## 3. METHODOLOGY

The core contribution of this work is a novel method for uncertainty estimation in the MVS stage, which is integrated into an uncertainty quantification framework for aerial and UAV photogrammetry. In the following subsections, **Section 3.1** presents the overall uncertainty quantification framework; **Section 3.2** briefly describes the standardized uncertainty estimation method in SfM; **Section 3.3** introduces our proposed method for uncertainty estimation in MVS.

### 3.1. Uncertainty Quantification Framework

The uncertainty quantification framework is adopted from the Generic Point-cloud Model (GPM) [93] developed by the National Geospatial Intelligence Agency (NGA). As shown in **Figure 2**, the uncertainty quantification framework propagates uncertainties from the SfM and MVS stages to the error covariance matrix of each 3D point in the dense point clouds:

$$\Sigma_{\mathrm{g}} = \left(\Sigma_{\varepsilon}^{-1} + \mathrm{B}_{\mathrm{X}}^{\mathrm{T}}\left(\mathrm{A}\Sigma_{\theta}\mathrm{A}^{\mathrm{T}} + \Sigma_{\mathrm{disp}}\right)^{-1}\mathrm{B}_{\mathrm{X}}\right)^{-1} \tag{1}$$

where $\Sigma_{\mathrm{g}}$ is the $3 \times 3$ error covariance matrix of a 3D point, including error propagation from the SfM and MVS stages. The framework treats their propagations additively, based on a practical decoupling of the dominant error sources inherent in the sequential reconstruction pipeline. The error propagation in the MVS stage assumes perfect camera poses, and this contribution is added to that of the SfM stage to obtain the final covariance. In reality, the impact of errors from SfM can be mitigated by searching dense correspondence in adjacent rows around the epipolar line. The uncertainty propagated from SfM captures the low-frequency, systematic errors that affect the entire model, while the uncertainty propagated from MVS represents the high-frequency, spatially varying errors that affect the local geometry. $\mathrm{B_X}$ is the $2n \times 3$ Jacobian matrix of $n$ 2D image observations with respect to the 3D point, modeled via the projection function. A is the $2n \times np$ Jacobian matrix of the same observations with respect to $np$ camera parameters modeled by the projection function. $\Sigma_\theta$ denotes the full $np \times np$ covariance matrix of the camera parameters obtained from $\Sigma_\mathrm{P}$ in BA, which will be discussed in **Section 3.2**. $\Sigma_{\mathrm{disp}}$ is the $2n \times 2n$ covariance matrix constructed from the disparity uncertainties of the $n$ matched pixels in the stereo pairs associated with the reference image, which will be discussed in **Section 3.3**. $\Sigma_\varepsilon^{-1}$ is a diagonal matrix with small positive values to stabilize the solution. The computed error covariance matrix for each 3D point is encoded in the dense point clouds in LAS format, in accordance with the specifications defined in the standardized framework [93].

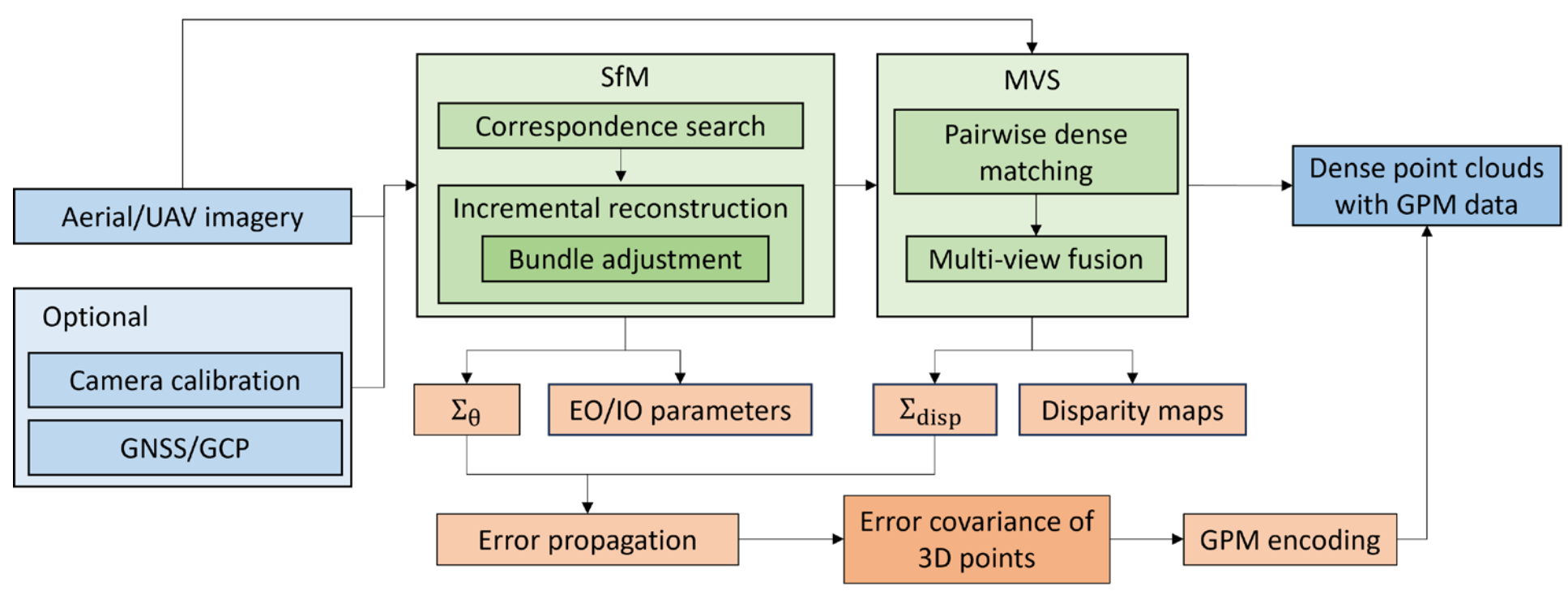

**Figure 2**
Overview of the uncertainty quantification framework.

In line with the GPM guidelines [93], our uncertainty quantification framework contains the information to compute the full covariance matrix for a set of sparsely sampled 3D points termed "anchor points" (AP):

$$\Sigma_{AP} = \begin{bmatrix} \Sigma_1 & \cdots & \Sigma_{1m} \\ \vdots & \ddots & \vdots \\ sym. & \cdots & \Sigma_m \end{bmatrix} \tag{2}$$

where $\Sigma_i$ is the error covariance matrix of the $i^{th}$ anchor point and $\Sigma_{ij}$ denotes the error cross covariance between the $i^{th}$ and $j^{th}$ anchor points. $\Sigma_{AP}$ is similarly derived from **Eq. 1**, where $\mathrm{B_X}$ becomes a $2n \times 3m$ Jacobian matrix of $n$ 2D image observations with respect to $m$ 3D point, modeled via the projection function, and A is the $2mn \times np$ Jacobian matrix. The full covariance matrix can be used to compute the relative error covariance between two anchor points:

$$\Sigma_{Rel_{i,j}} = \Sigma_i + \Sigma_j - \Sigma_{ij} - \Sigma_{ij}^T \tag{3}$$

It should be noted that we do not explicitly produce the full covariance matrix for dense point clouds in our experiments, as it would significantly increase computational costs and incur prohibitive storage costs. Following existing frameworks [90, 93], we compute per-point covariance to ensure a consistent comparison in our evaluation. This level of information is generally sufficient for most modern photogrammetric applications where dense point clouds or DSMs are treated as collections of individual 3D measurements. However, the framework retains the information required to construct the full covariance matrix, which can be computed if relative covariance analysis is needed. A detailed derivation of **Eq. 1** from the normal equations of multi-ray intersection under the GPM framework is provided in **Appendix 1**.

### 3.2. Uncertainty Estimation in SfM

We directly adopt established methods for uncertainty estimation in SfM [65], which have been extensively studied in Aerotriangulation [13, 32, 33]. In this subsection, we briefly outline the basic formulations of the problem. Readers are encouraged to refer to recent works [60, 64, 65] that provide more detailed discussions of efficient and numerically stable solutions.

At the core of SfM is BA, which optimizes the estimated camera parameters and 3D points to minimize the reprojection error:

$$\min_{X,\theta} \sum_{i,j} \left\| x_{ij} - \pi\left(\theta_i, X_j\right) \right\|^2 \tag{4}$$

where $X_j$ is the $j$-th 3D point, $\theta_i$ are the camera parameters of the $i$-th image, $x_{ij}$ is the observation of the $j$-th 3D point in the $i$-th image, and $\pi(\cdot)$ is the projection function. In the last iteration of BA, the system is approximated by linearizing the reprojection function:

$$r(P) \approx r(\hat{P}) + J(P - \hat{P}) \tag{5}$$

where $P = [X, \theta]$ stacks all parameters, r are the residuals, and J is the Jacobian matrix of residuals with regard to the parameters evaluated at $\hat{\theta}$ (the parameters at the last iteration). The covariance of the estimated parameters $\Sigma_P$ can be derived by:

$$\Sigma_P = (J^T \Sigma_x^{-1} J)^{-1} \tag{6}$$

where $\Sigma_x$ is the covariance matrix of the observations.

When available, our framework can be extended to the propagation of uncertainties associated with additional observations, such as those of GNSS, IMU and GCP. In this case, the resulting propagation reflects absolute uncertainty. These priors can be incorporated into the BA process by extending the objective function as:

$$\begin{gathered}\min_{X,\theta} \sum_{i,j} \left\| x_{ij} - \pi\left(\theta_i, X_j\right) \right\|^2 + \alpha \sum_{i} \left\| C_{i,\mathrm{pos}} - C_{i,\mathrm{GNSS}} \right\|^2 \\ + \beta \sum_{j} \left\| \log\left(R_{j,\mathrm{IMU}}^{-1} R_j\right) \right\|^2 + \gamma \sum_{k} \left\| X_k - X_{k,\mathrm{GCP}} \right\|^2\end{gathered} \tag{7}$$

where $C_{i,\mathrm{pos}}$ is the camera position of the $i$-th image associated with its GNSS data $C_{i,\mathrm{GNSS}}$, $R_j$ is the camera rotation of the $j$-th image associated with IMU-derived rotation $R_{j,\mathrm{IMU}}$, and $X_k$ is the $k$-th 3D point associated with GCP data $X_{k,\mathrm{GCP}}$. $\log(\cdot)$ is the matrix logarithm that maps a rotation matrix to a rotation vector. The weight factors $\alpha$, $\beta$ and $\gamma$ are chosen inversely proportional to the expected nominal accuracy value of each observation type [94, 95]. For general aerial and UAV datasets, initial values of the weights can be set as $\alpha = 1 - 100$ for GNSS, $\beta = 0.1 - 10$ for IMU, and $\gamma = 100 - 1000$ for high-accuracy GCP. These weights can be further refined if accuracy statistics are available for the specific dataset. We can then construct the residuals accordingly and extend the Jacobian matrix as:

$$J_{total} = \begin{bmatrix} J \\ J_{\mathrm{GNSS}} \\ J_{\mathrm{IMU}} \\ J_{\mathrm{GCP}} \end{bmatrix} \tag{8}$$

which corresponds to the covariance matrix of all available observations:

$$\Sigma_{\mathrm{obs}} = \begin{bmatrix} \Sigma_x & 0 & 0 & 0 \\ 0 & \Sigma_{\mathrm{GNSS}} & 0 & 0 \\ 0 & 0 & \Sigma_{\mathrm{IMU}} & 0 \\ 0 & 0 & 0 & \Sigma_{\mathrm{GCP}} \end{bmatrix} \tag{9}$$

Finally, the covariance of the estimated parameters $\Sigma_P$ is given by:

$$\Sigma_P = \left(J_{\mathrm{total}}^T \Sigma_{\mathrm{obs}}^{-1} J_{\mathrm{total}}\right)^{-1} \tag{10}$$

### 3.3. Uncertainty Estimation in MVS

We employ a robust MVS method [96-98] for dense reconstruction, which has been widely applied in aerial and UAV photogrammetry. The method uses Census-based [99] Semi-Global Matching (SGM) [100] for stereo matching. Specifically, each image in the SfM reconstruction is treated as a reference view and paired with a set of neighboring views to form stereo pairs. Each stereo pair undergoes image rectification and stereo matching, resulting in multiple disparity maps per reference view. These disparity maps are then converted into depth maps. The depth maps are refined using median filtering [101] to remove outliers, yielding one fused depth map per reference view. A 3D point is triangulated only if it is observed in at least three views. All per-view point clouds are merged as a unified 3D reconstruction. It should be noted that our uncertainty quantification framework is conceptually adaptable to other MVS methods [66] as long as similar inputs (correspondences and uncertainty) are given for error propagation.

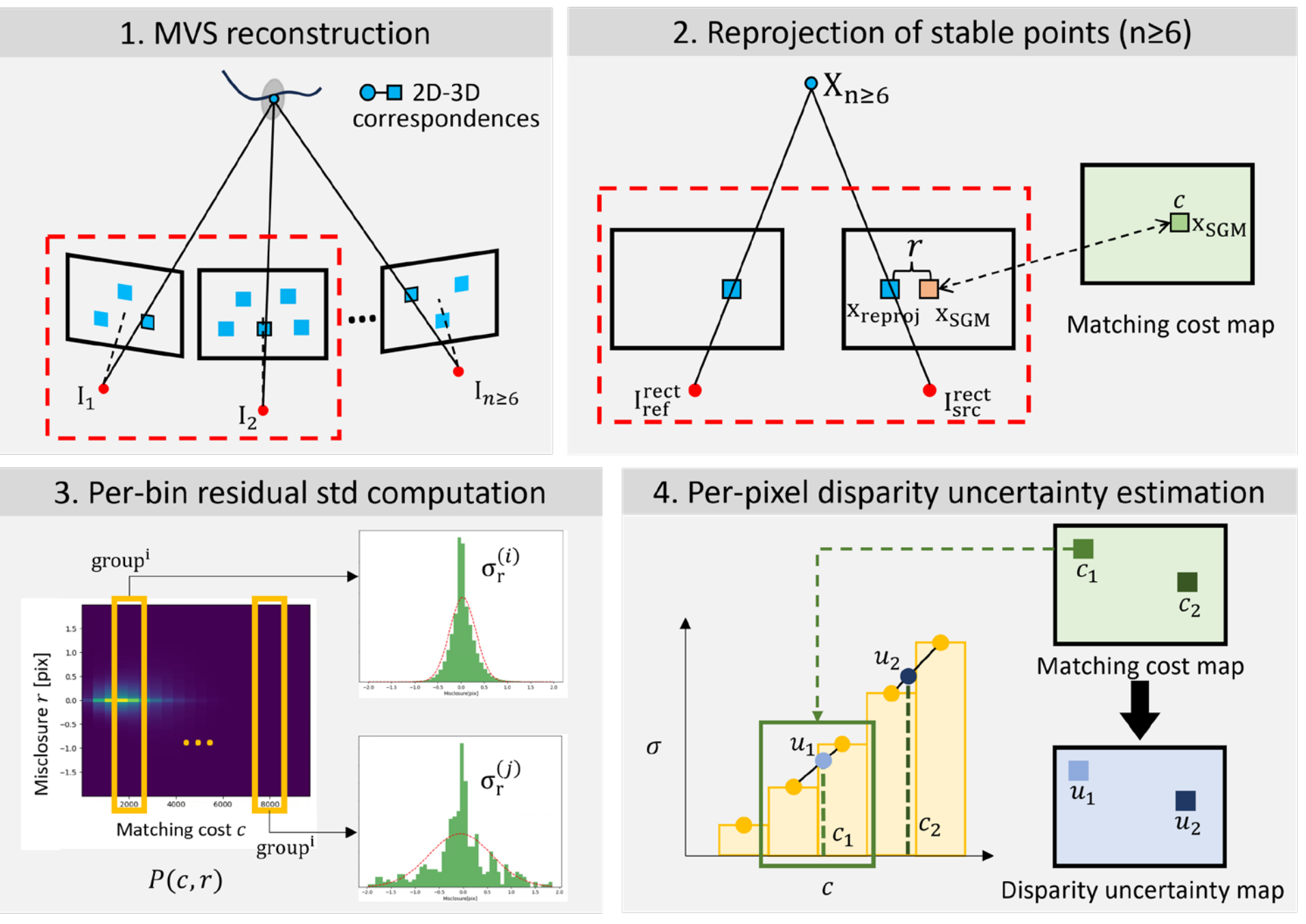

**Figure 3**

Overview of our method of uncertainty estimation in the MVS stage. Step 1: n-view points (n≥6) are selected from the point clouds. Step 2: Each n-view point is reprojected to every stereo pair to compute the residual $r$, associated with the matching cost $c$. Step 3: Each stereo pair constructs a joint distribution of $c$ and $r$, of which the standard deviation $\sigma$ of $r$ is computed for each bin with discrete intervals of $c$. Step 4: The disparity uncertainty $u$ for each pixel is interpolated between two consecutive $c$-$\sigma$ bins according to its matching cost $c$. Finally, the disparity uncertainties from multiple stereo pairs constitute the uncertainty in the MVS stage.

The objective of the proposed method is to estimate uncertainty in the aforementioned MVS process for each reference view, which encompasses disparity uncertainties across all stereo pairs, i.e., each pixel in a pairwise disparity map is associated with an uncertainty value. Our preliminary study [48] yields the following findings: 1) the number of views is a strong indicator of point accuracy; 2) by separating points into high- and low-accuracy groups, the distribution of n-view points clearly distinguishes reliable points, with a threshold around n=6; 3) such n-view points (n≥6) constitute approximately 30% of all points, providing a sufficient set of stable points; and 4) the matching cost is a highly informative cue that correlates with stereo matching quality, but its scale varies across different stereo pairs. Therefore, we leverage the n-view points (n≥6) as pseudo-check points to: 1) regress disparity uncertainty using matching cost for all points, and 2) self-calibrate the magnitude of disparity uncertainty across different stereo pairs. **Figure 3** provides an overview of our proposed method. It begins by reprojecting each 3D point to every rectified stereo pair, and the residual $r$ along the epipolar direction is computed as:

$$\begin{aligned} \mathrm{r} &= \mathrm{x}_{\mathrm{reproj}} - \mathrm{x}_{\mathrm{SGM}} \\ \mathrm{x}_{\mathrm{reproj}} &= \pi(\mathrm{P}, \mathrm{X}) \end{aligned} \tag{11}$$

where $\pi$ is the projection function that maps the 3D point X to the rectified source view using its camera parameters P. $\mathrm{x}_{\mathrm{SGM}}$ denotes the dense correspondence obtained from the SGM algorithm, represented as sub-pixel estimates. The value of r can be either positive or negative, depending on the relative position of the reprojected pixel $\mathrm{x}_{\mathrm{reproj}}$ and the dense correspondence $\mathrm{x}_{\mathrm{SGM}}$. Additionally, the matching cost c at $\mathrm{x}_{\mathrm{SGM}}$ is retrieved from the precomputed matching cost map. For each stereo pair, all n-view points (n≥6) are then selected to obtain corresponding r and c values, which are used to construct a joint distribution $\mathrm{P}(\mathrm{c}, \mathrm{r})$. Next, these n-view points (n≥6) are grouped into discrete bins of equal interval based on increasing matching cost c. The number of bins is set to 8, based on the typical range of matching costs in the SGM algorithm. The bin width is adaptively defined as $\frac{c_{99.7}-c_{min}}{8}$, where $c_{99.7}$ corresponds to the 99.7th percentile (i.e., the 3σ level) to exclude outliers. We model the conditional distribution of the residual r given that the matching cost c falls within the $k$-th bin as $\mathrm{P}(\mathrm{r}|\mathrm{c} \in \mathrm{bin}_k)$, and compute the standard deviation $\sigma_{\mathrm{r}}^{(k)}$ of r within the $k$-th bin as:

$$\sigma_{\mathrm{r}}^{(k)} = \sqrt{\frac{1}{\mathrm{N}_k - 1}\sum_{i=1}^{\mathrm{N}_k}\left(\mathrm{r}_i^{(k)} - \bar{\mathrm{r}}^{(k)}\right)^2} \tag{12}$$

where $\mathrm{N}_k$ is the number of points in the $k$-th bin, $\mathrm{r}_i^{(k)}$ are the residual values, and $\bar{\mathrm{r}}^{(k)}$ is the mean residual in the $k$-th bin. A minimum of 5 points is required; otherwise, the number of bins is reduced by one, and the conditional distribution is recomputed, repeating this process until the requirement is satisfied. Each stereo pair maintains a set of c−σ bins to regress per-pixel disparity uncertainty. Specifically, for each pixel in the disparity map, the uncertainty value u is obtained by linearly interpolating between the σ values of the two bins whose c values are closest to the pixel's matching cost. For pixels that are triangulated into 3D points, the uncertainty value u is further refined using the residual r as follows:

$$\mathrm{u} = \frac{\max(\mathrm{u}, |\mathrm{r}|) + \mathrm{u}}{2} \tag{13}$$

This refinement preserves the uncertainty value u when it is larger than the residual, but increases the uncertainty value u when the residual exceeds it, providing a smooth transition between the two values. Finally, the disparity uncertainties from all stereo pairs constitute the uncertainty in MVS for each reference view. Each pixel in the reference view is associated with $n$ disparity uncertainty values, corresponding to the number of matched pixels from neighboring views. Each disparity uncertainty, as defined in the epipolar space, is projected back to the original image space and decomposed into x and y components. These components constitute the $\Sigma_{\mathrm{disp}}$ as described in **Section 3.1**, which is propagated through **Eq. 1** to obtain the covariance matrix of each 3D point.

The key innovation of our method is the adaptation of algorithm-generated measurements with estimated uncertainty in MVS to obtain the final covariance matrix of 3D points (from both SfM and MVS). Our method offers several key advantages over existing approaches. First, it does not require any training data. The use of self-contained n-view points (n≥6) within MVS enables self-calibrating disparity uncertainty magnitudes across different stereo pairs. Second, the uncertainty estimation is continuous, rather than yielding discrete outputs as in TV-based methods [47, 102] or relying on fixed, hard-coded value [103, 104], allowing for finer granularity. Third, the approach is computationally efficient and scalable to large-scale airborne and UAV imagery, as it avoids the use of memory-intensive cost volumes commonly employed in existing DL-based methods [45, 46, 84, 86]. Finally, by leveraging the multi-view information inherent in MVS, the method effectively identifies large uncertainties that may not be directly captured by pairwise matching costs alone.

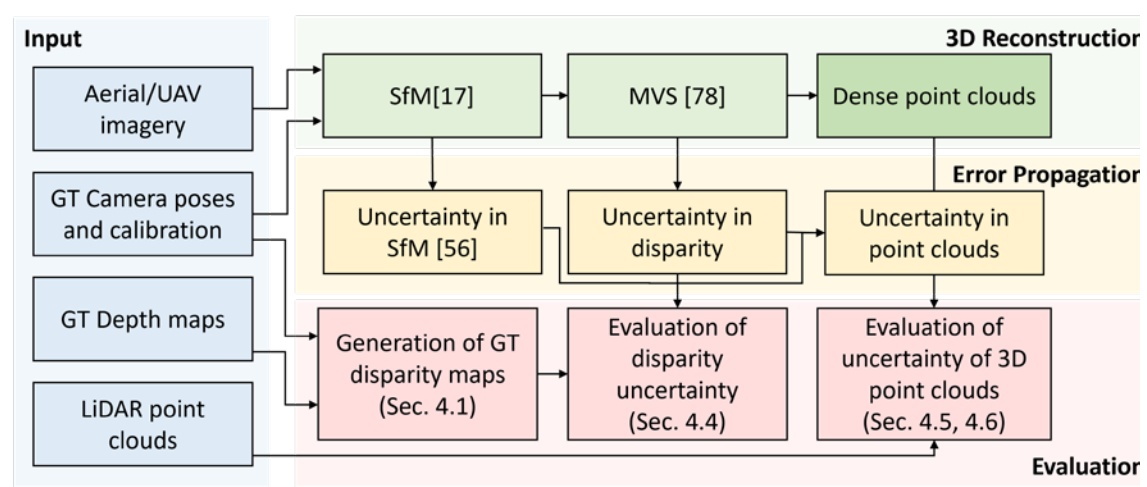

**Figure 4**
Overview of the data processing procedure in our evaluation.

## 4. EXPERIMENTAL RESULTS

The existing works [46, 47, 90, 92] focus on the evaluation of either disparity or 3D point clouds, lacking a comprehensive assessment of both, particularly with respect to uncertainty estimation – including overall uncertainty quantification and the level of trustworthiness of the predicted uncertainty bound. Therefore, we propose a new evaluation metric dedicated to this purpose and conduct comprehensive experiments using publicly available airborne and UAV datasets to evaluate the uncertainty quantification framework. The evaluation covers both disparity uncertainty and the uncertainty of 3D point clouds, the outline of the evaluation procedure is shown in **Figure 4**. **Section 4.1** describes the evaluation datasets and the preparation of GT disparity maps. The evaluation setup and metrics are detailed in **Section 4.2** and **Section 4.3**, respectively. Quantitative and qualitative results, along with analysis of disparity uncertainty, the relative uncertainty of 3D point clouds, and the absolute uncertainty of 3D point clouds, are presented in

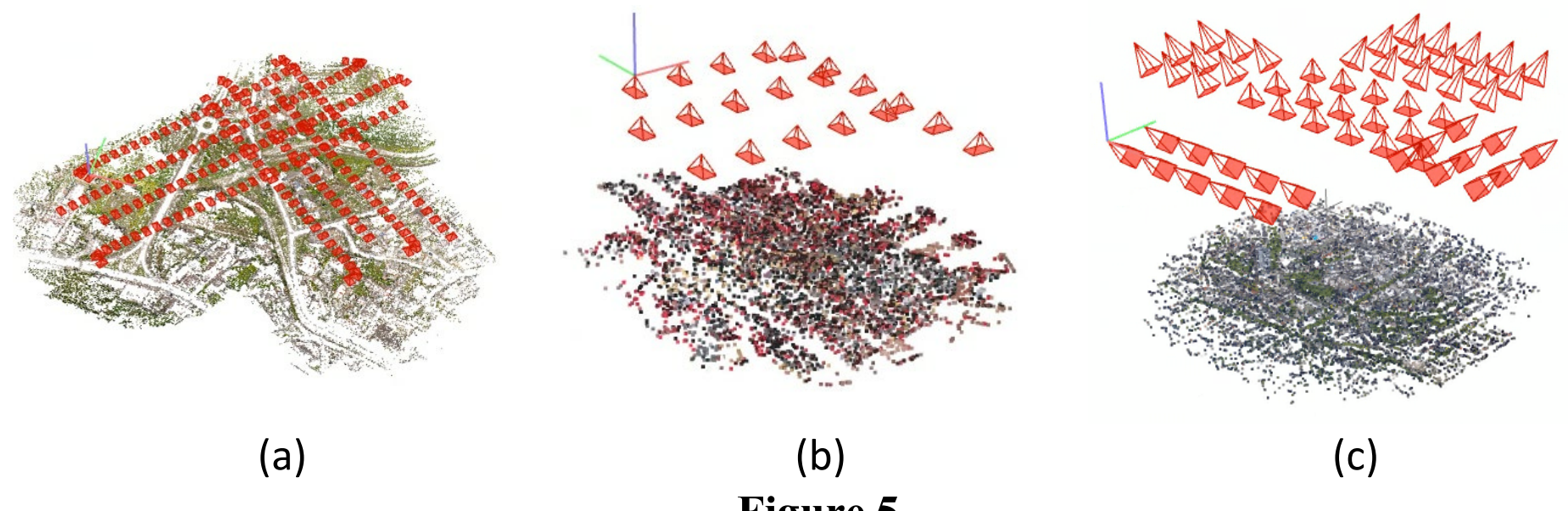

**Figure 5**

Overview of UAV dataset (a) UseGeo and airborne datasets (b) Vaihingen, (c) Dortmund.

**Table 1**

Specification of the datasets. For the Dortmund dataset, the focal lengths and overlaps are listed in the order <nadir, oblique>.

| | UseGeo | Vaihingen | Dortmund |
|---|---|---|---|
| # images | 829 | 20 | 59 |
| GSD [cm] | 1.7 –1.9 | 8.0 | 8.0 – 12.0 |
| Image size | 1989 x 1320 | 13824 x 7680 | 6132 x 8176 |
| GT camera poses and calibration | Y | Y | N |
| GT depth maps or LiDAR data | Y | Y | Y |
| Flying height [m] | 80 | 900 | 905 |
| Focal length [pix] | 1160 | 10000 | 8333, 13667 |
| Overlap (forward/side) [%] | 80/60 | 60/60 | 75/80, 80/80 |

**Sections 4.4**, **Section 4.5** and **Section 4.6. Section 4.7** provides a sensitivity analysis of the choice of n-view points.

### 4.1. Datasets and GT Disparity Preparation

Three airborne and UAV datasets are used for evaluation. Visualization and specification of each dataset are provided in **Figure 5** and **Table 1,** respectively. The datasets and the preparation of GT disparity data are introduced as follows.

**UseGeo**. UseGeo [105, 106] is a UAV-based multi-sensor dataset for multiple geospatial tasks such as monocular depth estimation and multi-view 3D reconstruction. Three flights are performed to collect data on three different urban and peri-urban areas in Italy, respectively with 224 images, 328 images, and 277 images for each sub-dataset. All images are captured at nadir angles. Images with GT camera poses and calibration are provided. Additionally, dense GT depth maps are available. It is feasible to assess both disparity uncertainty and uncertainty of 3D point clouds.

**Vaihingen**. Vaihingen airborne dataset [107] is part of ISPRS benchmark on urban object classification and 3D building reconstruction [108]. It contains 20 color infrared images that cover three different types of areas from nadir views: "inner city" with historic buildings with complex shapes, "high riser" with a few high-rising residential buildings, and "residential area" with small and detached houses. Similar to UseGeo, the Vaihingen dataset also provides GT camera poses, calibration, and airborne LiDAR data. Therefore, it is also used for our evaluation on both disparity uncertainty and uncertainty of 3D point clouds.

**Dortmund**. Dortmund airborne dataset [109] is part of ISPRS/EuroSDR benchmark for multi-platform photogrammetry. The subset contains 59 nadir and oblique images captured using the airborne imaging system (IGI PentaCam) with five camera heads [110]. Airborne LiDAR data is available for evaluation on the uncertainty of 3D point clouds in our experiments. However, the evaluation of disparity uncertainty is infeasible due to the lack of GT poses. Additionally, the evaluation of absolute uncertainty is also infeasible due to the lack of georeferencing information.

For UseGeo and Vaihingen datasets, GT depth maps, camera poses, and calibration are available. Therefore, the GT disparity maps can be derived. First, the stereo pair is rectified using GT camera poses and calibration.

Next, the GT depth maps are converted to align with the rectified images. The GT disparity maps can be generated using the following equation:

$$d = \frac{b * f}{D} \tag{14}$$

where b is the baseline, f is the focal length of the rectified images, and D is the depth. For Vaihingen dataset, an extra step is performed to convert the LiDAR-derived DSM provided by the dataset to the depth maps before generating the GT disparity maps.

The GT disparity maps are further refined by checking left-right disparity consistency. Two disparity maps from the same stereo pair with image orders reversed are compared. Disparity values with a difference greater than one pixel between the two maps are filtered out. This post-processing step excludes the outliers and occlusions in the GT disparity maps.

### 4.2. Experiment Setup

In our experiment, the photogrammetric point clouds are generated using a reconstruction pipeline consisting of the widely used SfM package Colmap [15], combined with a robust MVS method [96, 98] based on SGM. The data processing for each evaluation task is described in detail as follows.

**Evaluation of disparity uncertainty**. Disparity uncertainty is a key component of our proposed method for uncertainty estimation in the MVS stage. Therefore, we conduct dedicated experiments to evaluate the disparity uncertainty independently, isolating it from the complex error propagation in the subsequent step. UseGeo and Vaihingen datasets are used to evaluate the performance of disparity uncertainty estimation. For each UseGeo sub-dataset, we select 5 reference images, which are evenly distributed in the area. Each reference image is matched to 8 neighboring views, constructing 120 stereo pairs in total. For Vaihingen dataset, we select 2 reference images covering the most area where LiDAR data is available for evaluation. Each reference image is matched to 10 neighboring views to build 20 stereo pairs. We fix the camera poses and calibration provided by GT to evaluate the disparity uncertainty produced by the stereo matching algorithm alone.

Our method is compared with two existing methods: the TV-based method [47, 83] and SEDNet [46]. The TV-based method is applied in the existing uncertainty quantification framework [90], while SEDNet represents a recent DL-based approach. Other DL-based approaches [45, 87] are excluded since we have no access to their code. We use SGM as the stereo matching algorithm for the TV-based method and our proposed method. We follow the equations described in [47, 83] to compute TV classes and use the provided lookup table to map the TV classes to disparity uncertainty. For SEDNet, we use the pretrained models for evaluation rather than training the network ourselves for several reasons. First, the cost volume aggregation in SEDNet is based on a limited disparity range starting from zero. While this design is suitable for synthetic or autonomous driving datasets with small baselines, it is not applicable to large-scale airborne and UAV datasets, which often exhibit arbitrary and much wider disparity ranges. Second, the resolution of airborne and UAV imagery is several orders of magnitude higher than the datasets used for SEDNet, resulting in GPU memory constraints during training. Since the disparity prediction network and the uncertainty estimation module in SEDNet are tightly coupled, the entangled prediction of disparity and uncertainty may inevitably create inherent correlations. For example, large disparity values may come with low uncertainties, and vice versa. Thus, to decorrelate the uncertainty from disparity in SEDNet and enable a fair comparison with our method, we evaluate the disparity uncertainty only at pixels where SEDNet shares the same disparity values as SGM (with a difference of less than one pixel). This ensures that the comparison reflects the uncertainty values from both methods under the condition of the same measurement (disparity). On the other hand, the comparison between our method and the TV-based method uses the same disparity maps and assesses over all pixels. The estimated uncertainty is evaluated against the actual error between disparity values from the stereo matching algorithm and the GT.

**Evaluation of uncertainty of 3D point clouds**. All datasets are used to evaluate the uncertainty of 3D point clouds. For the Dortmund dataset, 5 evenly distributed reference images are selected, each is matched with 10 neighboring views. 3D point clouds are generated for the selected reference images in all datasets. Unlike the evaluation of disparity uncertainty, camera poses and calibration are optimized through BA to incorporate the uncertainty in SfM.

We compare the uncertainty quantification framework with the one in [90], which is the only existing framework that propagates uncertainty through the photogrammetric reconstruction processes by integrating the TV-based method. Since the code for [90] is unavailable, we compare the uncertainty quantification framework integrated with either TV-based method or our proposed method. Uncertainty from the SfM process is obtained using the existing framework [65]. For evaluation, the dense 3D point clouds are aligned to LiDAR point clouds using the Iterative Closest Point (ICP) algorithm [111] to eliminate systematic errors. Since point-wise correspondences between the reconstruction and LiDAR point clouds are unknown, the actual error of each reconstructed 3D point is measured as the distance to a quadratic plane fitted by the closest 6 points from the LiDAR point clouds for improved robustness [112-114], and we compare the radius of sphere spanned by the estimated covariance matrix with the actual error for each point. For evaluation of absolute uncertainty, we compare the 3D point clouds directly with the LiDAR point clouds without ICP alignment.

**Table 2**
Quantitative evaluation of disparity uncertainty. For each comparison, better results are highlighted in bold.

| | | | # pixels | Pearson coefficient ↑ | Mean [px] ↓ | RMSE [px] ↓ | KL divergence ↓ | Bounding rate [%] ↑ |
|---|---|---|---|---|---|---|---|---|
| UseGeo | SGM | TV-based | 186.3M | 0.191 | 1.155 | 3.324 | 1.153 | **65.373** |
| | | Ours | | **0.472** | **0.778** | **2.608** | **0.794** | 65.327 |
| | DL | SEDNet | 129.7M | 0.146 | 3.027 | 3.667 | 0.895 | **98.306** |
| | | Ours | | **0.279** | **0.547** | **1.848** | **0.775** | 68.685 |
| Vaihingen | SGM | TV-based | 17.7M | 0.138 | 3.413 | 9.352 | 1.609 | **56.568** |
| | | Ours | | **0.226** | **3.147** | **9.302** | **1.427** | 48.444 |
| | DL | SEDNet | 2.8M | 0.060 | 12.650 | 13.845 | 1.597 | **96.891** |
| | | Ours | | **0.169** | **1.680** | **6.308** | **1.499** | 50.921 |

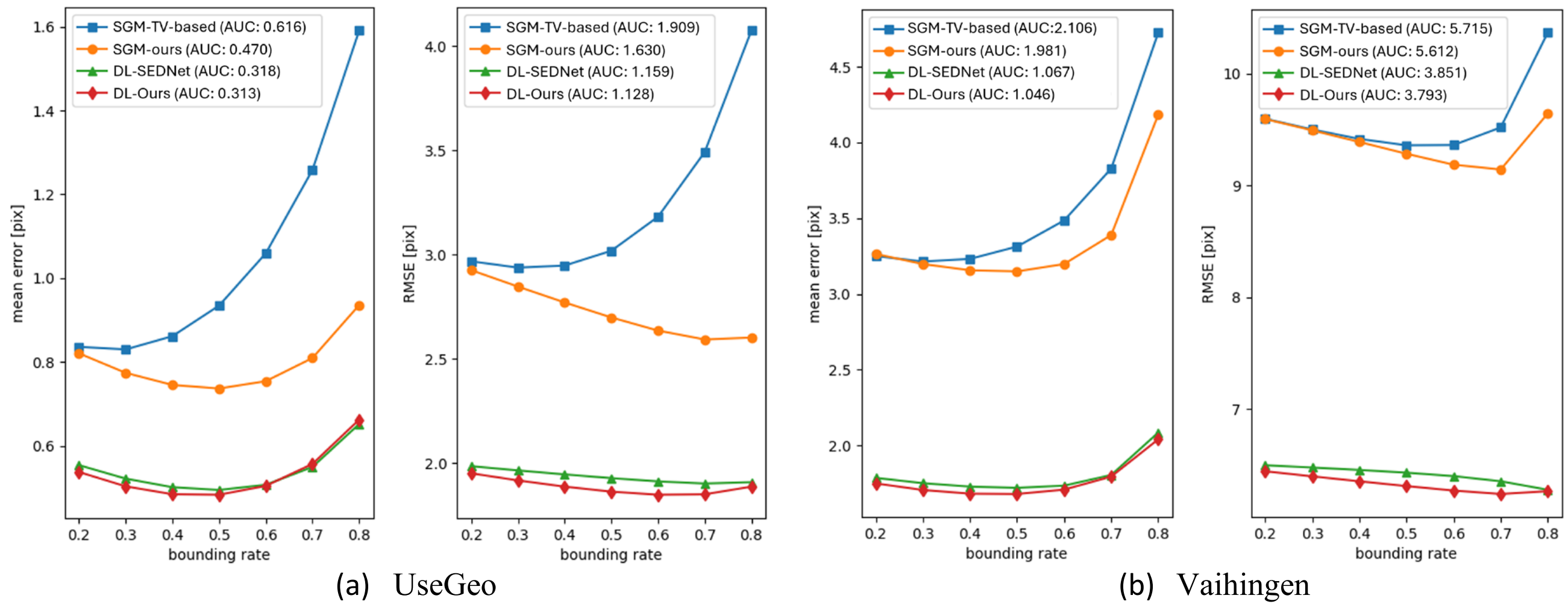

(a) UseGeo (b) Vaihingen

**Figure 6**
Accuracy performance under different bounding rates.

### 4.3. Evaluation Metrics

We use several common metrics for a comprehensive assessment of both the disparity uncertainty and the uncertainty of 3D point clouds. Pearson correlation coefficient is used to evaluate the linear relationship between the predicted uncertainty and the actual errors. A higher Pearson coefficient is favored. KL divergence is another metric to measure how the distribution of predicted uncertainty matches the distribution of the actual errors. Lower KL divergence represents a better alignment between the distributions. Mean error and RMSE are two metrics to indicate the accuracy. Since existing metrics do not fully reflect whether the estimated uncertainty is meaningful and certifiable (e.g., cases where the predicted uncertainty of disparity or 3D point is accurate but consistently lower than the actual error), we propose bounding rate $\rho_{BR}$ as a complement metric to reflect how well the predicted uncertainty bounds the actual errors:

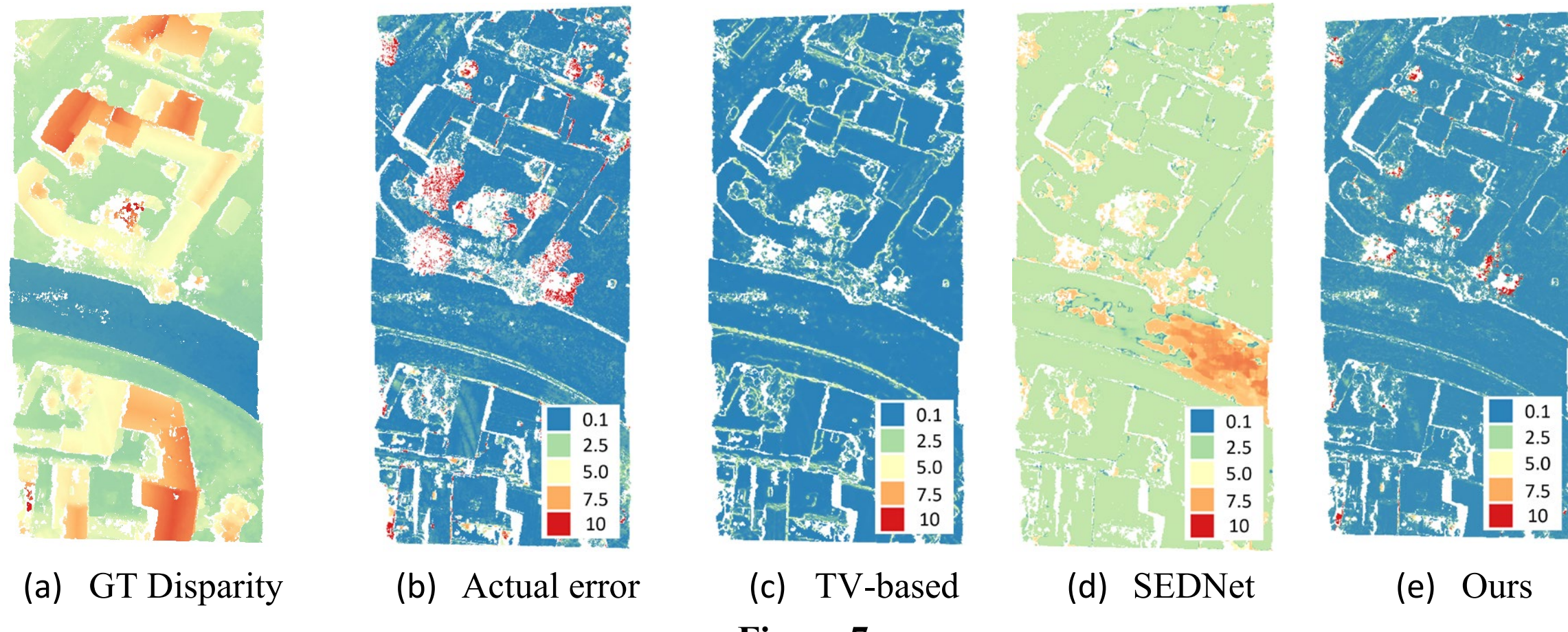

**Figure 7**

Visualization results of disparity uncertainty estimation on UseGeo dataset. The predicted uncertainty (in pixels) by different methods are displayed in the same color scale.

$$\rho_{\mathrm{BR}} = \frac{\sum_{i=1}^{N} 1(\mathrm{actual} \leq \mathrm{pred})}{N} \tag{15}$$

Where $1(\cdot)$ is the indicator function. $\rho_{\mathrm{BR}}$ refers to the percentage of pixels or 3D points whose actual errors are within the range of predicted uncertainty. It reflects the fidelity of the uncertainty estimates and serves as a truncated metric for estimating the trustworthiness of the predicted uncertainty bound. While bounding rate and accuracy are distinct metrics, they enable the evaluation of uncertainty from different perspectives. When combined, they can indicate if the uncertainty is overestimated or not. Ideally, a tighter bound of uncertainty is favored to balance the bounding rate and accuracy. To further compare the methods, the accuracy under different bounding rates by scaling the prediction of uncertainty is evaluated. Area under Curve (AUC) for accuracy (mean error or RMSE) against different bounding rates is evaluated, and lower AUCs are favored.

### 4.4. Evaluation of Disparity Uncertainty

We first compare our method with the TV-based approach and SEDNet in estimating disparity uncertainty. **Table 2** presents the quantitative results on the UseGeo and Vaihingen datasets. The results demonstrate that the uncertainty estimated by our method more closely reflects the actual error compared to other approaches. When compared to the TV-based method over all pixels, our approach achieves superior performance across most evaluation metrics, including correlation, accuracy, and distribution alignment. Specifically, our method shows up to 147.1% improvement in Pearson correlation, 32.6% reduction in mean error, 21.5% reduction in RMSE, and 31.0% reduction in KL divergence. Although the TV-based method achieves a higher bounding rate by assigning discrete uncertainty levels—often larger than the actual error—this comes at the cost of lower granularity, as reflected in the other metrics. In comparison to SEDNet over the subset of pixels as described in **Section 4.2**, our method shows even more pronounced advantages. While SEDNet achieves significantly higher bounding rates, which bounds the actual errors on up to 1.9 times more pixels than our method, it shows a dramatic performance drop in the other metrics. Our method achieves up to 181.7% improvement in Pearson correlation, 86.7% reduction in mean error, 54.3% reduction in RMSE, and 13.4% reduction in KL divergence. These results suggest that SEDNet tends to overestimate disparity uncertainty. However, it should be noted that the comparison with SEDNet is limited, as the evaluation is restricted to pixels where SEDNet and SGM produce similar disparities. This filtering ensures a fairer comparison of uncertainty estimates, but introduces a bias toward less challenging regions. For example, object edges, which are expected to exhibit large disparity errors, are often smoothed out because SEDNet downsamples the input images. As a result, the performance of SEDNet in these regions remains underexplored.

Further evaluation is performed to assess the accuracy performance under varying bounding rates by scaling the predicted disparity uncertainty accordingly for each method. “Ours” is compared with the TV-based method over all pixels, while “Ours sampled” is compared with SEDNet over the subset of pixels as described in **Section 4.2**. **Figure 6** illustrates that achieving high

**Table 3**

Quantitative evaluation of uncertainty of 3D point clouds. For each dataset, better results are highlighted in bold.

| | | # points | Pearson coefficient ↑ | Mean [m] ↓ | RMSE [m] ↓ | KL divergence ↓ | Bounding rate [%] ↑ |
|---|---|---|---|---|---|---|---|
| UseGeo | TV-based | 4.1M | 0.174 | **0.062** | 0.109 | 0.748 | 55.988 |
| | Ours | | **0.198** | **0.062** | **0.106** | **0.669** | **61.730** |
| Vaihingen | TV-based | 14.2M | 0.206 | 0.101 | 0.241 | 0.641 | 39.151 |
| | Ours | | **0.249** | **0.099** | **0.240** | **0.602** | **40.908** |
| Dortmund | TV-based | 21.8M | **0.133** | **0.249** | 0.637 | **0.858** | 33.857 |
| | Ours | | 0.123 | 0.254 | **0.627** | 0.892 | **51.794** |

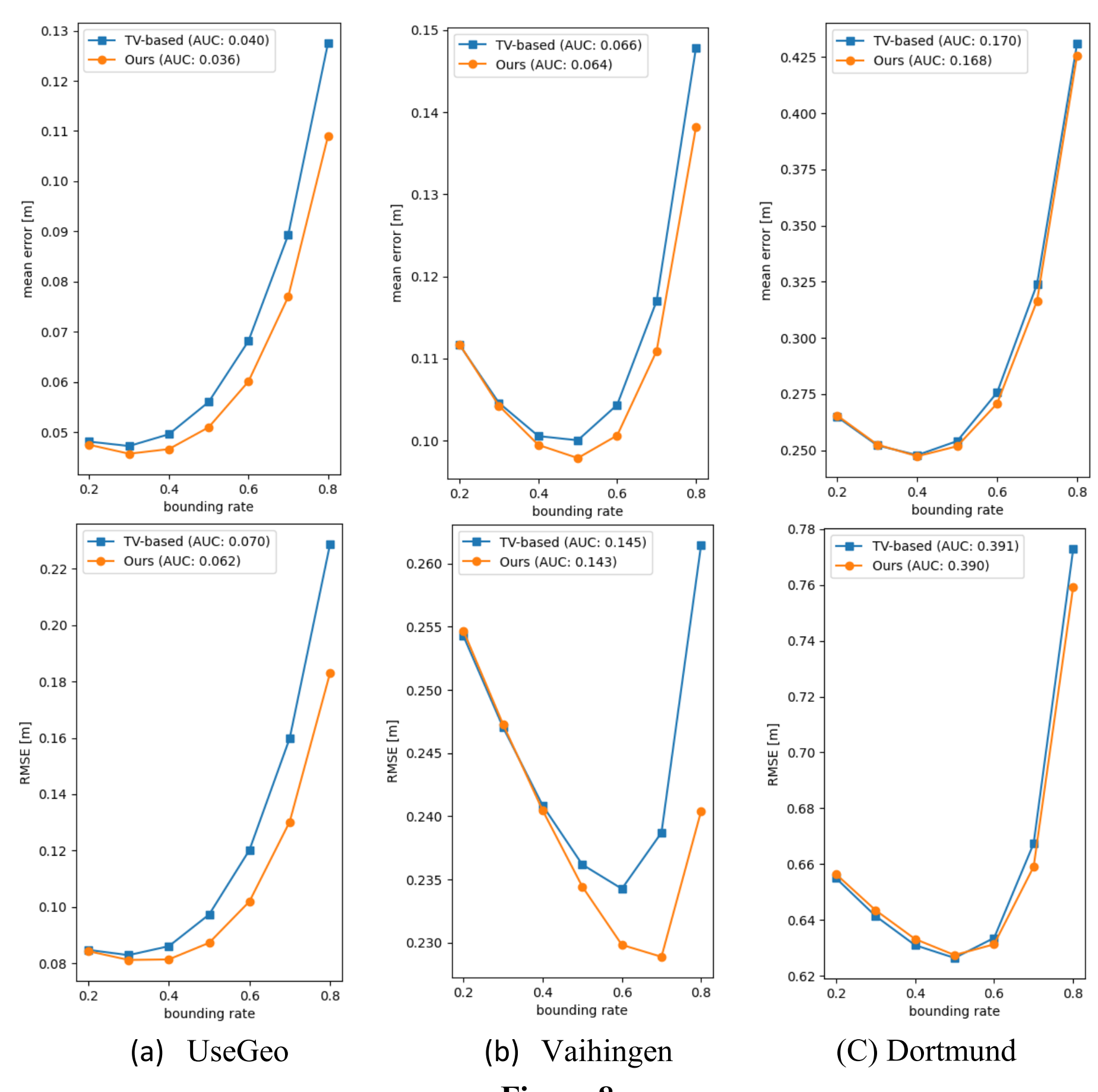

(a) UseGeo (b) Vaihingen (C) Dortmund

**Figure 8**

Accuracy performance under different bounding rates. First row: mean error. Second row: RMSE.

bounding rates generally leads to higher mean error and RMSE, indicating a trade-off between bounding rates and accuracy. Our method consistently outperforms the TV-based method across all bounding rates, with the performance gap widening as the bounding rate increases. In terms of AUC, our method achieves up to 23.70% improvement in mean error and 14.61% improvement in RMSE over the TV-based approach, which verifies its superior accuracy in tightly bounding actual disparity errors. Compared to SEDNet, our method demonstrates similar performance when evaluated on the subset of pixels ("Ours sampled"). Both methods exhibit significantly lower errors than those observed in the full-pixel evaluation. The reason is that the subset of pixels is mostly distributed on large planar surfaces, where disparity errors are typically smaller and more predictable. In contrast, other challenging regions such as object boundaries and fine structures with large errors are excluded from comparison due to inconsistent disparity values between

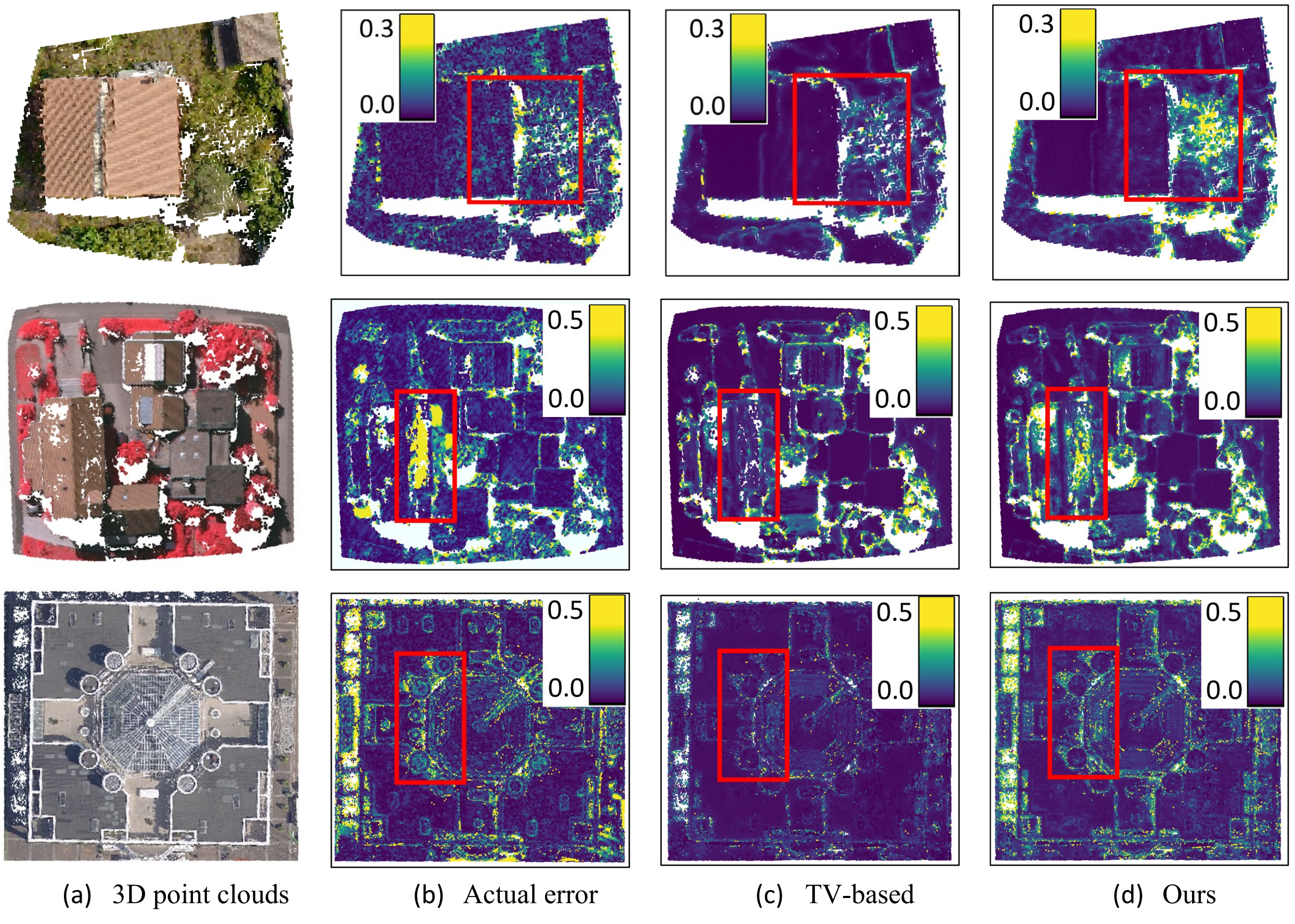

**Figure 9**

Visualization of uncertainty of 3D point clouds with compared to the actual errors. First row: UseGeo. Second row: Vaihingen. Third row: Dortmund. Color scale is in meters.

SEDNet and SGM. SEDNet has poor disparity prediction in the excluded region, as these fine details are smoothed out during image downsampling operations in SEDNet. At $c_{99.7}$, the proposed method achieves a bounding rate of approximately 70%, indicating a balanced trade-off between coverage and the tightness of the uncertainty bounds. In contrast, SEDNet tends to achieve near 100% coverage, at the expense of more conservative uncertainty estimates.

**Figure 7** presents a visual comparison of the predicted disparity uncertainties from different methods. TV-based approach produces discrete disparity uncertainty levels, with the maximum uncertainty capped by the lookup table. This limitation highlights its poor generalization, as the method is trained solely on an indoor dataset and fails to adapt to more diverse scenes. SEDNet consistently overestimates disparity uncertainty across most pixels. Although this leads to a higher bounding rate, it comes at the expense of significant accuracy degradation. On the other hand, our method generates continuous uncertainty estimates and effectively identifies regions with genuinely high uncertainty, e.g., trees, while maintaining tighter bounds on actual errors in the remaining regions. Overall, our method demonstrates superior performance by achieving a more favorable trade-off between accuracy and bounding rates.

### 4.5. Evaluation of Relative Uncertainty of 3D point clouds

We evaluate the uncertainty quantification framework when integrated with either the TV-based method or our proposed method. As shown in **Table 3**, the uncertainty quantification framework with our proposed method consistently outperforms the framework with TV-based method on the UseGeo and Vaihingen datasets across all evaluation metrics. Specifically, our method demonstrates up to 20.9% improvement in Pearson correlation, 2.0% reduction in mean error, 2.8% reduction in RMSE, 10.6% reduction in KL divergence, and 10.3% in bonding rate. For the Dortmund dataset, the performance of the two methods is comparable in

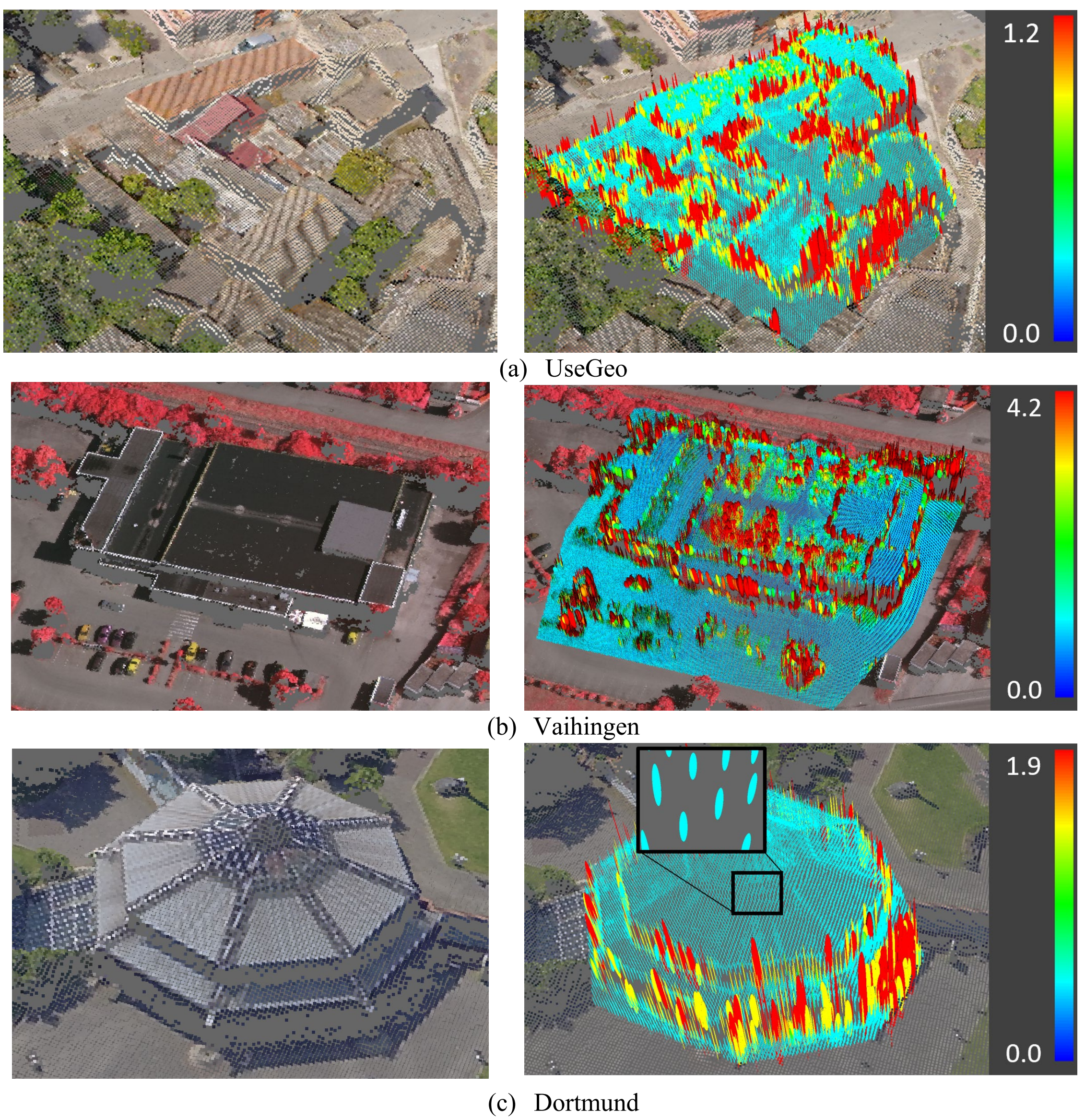

(a) UseGeo

(b) Vaihingen

(c) Dortmund

**Figure 10**

Visualization results of 3D covariance matrices of point clouds. First column: photogrammetric dense point clouds. Second column: visualization of covariance metrices in ellipsoids. The ellipsoids are slightly enlarged for better visualization, while their actual size (in meters) is indicated by the color scale.

most metrics. However, our method achieves a significantly higher bounding rate with 53.0% of improvement. Overall, the performance gap of uncertainty of 3D point clouds is less pronounced than that of disparity uncertainty due to the error propagation process. In most cases, the disparity uncertainties predicted by the two methods differ by only one to two pixels. As these uncertainties are propagated to 3D points, the uncertainties become more similar. **Figure 8** further illustrates the accuracy comparison under varying bounding rates. Uncertainty quantification framework integrated with our method achieves lower mean error and RMSE as the predicted uncertainties are scaled to achieve higher bounding rates on the UseGeo and Vaihingen datasets. This demonstrates our method's robustness in maintaining accuracy even when scaling to bound more pixels. For the Dortmund dataset, the performance curves of both methods are closely aligned, which is consistent with the trends observed in the quantitative evaluation. Our method achieves up to 10.0% improvement in AUC for mean error and up to 11.43% improvement in AUC for RMSE, which indicates its superior performance in balancing accuracy and bounding rates.

**Table 4**
Quantitative evaluation of absolute uncertainty of 3D point clouds

| | # points | Pearson coefficient | Mean [m] | RMSE [m] | KL divergence | Bounding rate [%] |
|---|---|---|---|---|---|---|
| UseGeo | 4.1M | 0.201 | 0.071 | 0.118 | 0.544 | 67.960 |
| Vaihingen | 14.2M | 0.132 | 0.144 | 0.341 | 0.601 | 57.503 |

**Figure 9** provides a visual comparison of the uncertainty of 3D point clouds estimated by the uncertainty quantification framework integrated with different methods for uncertainty estimation in MVS. Our method effectively captures high uncertainty in challenging regions such as trees, small objects, and building boundaries, as highlighted by the red boxes in the first and third rows. It also accurately reflects increased uncertainty in textureless areas or regions with repetitive patterns such as rooftops, shown in the red boxes of the first and second rows. In contrast, uncertainty quantification framework with TV-based method fails to identify these high-uncertainty areas. This difference stems from our method's ability to leverage multi-view points to refine disparity uncertainty, which successfully identifies the regions with high uncertainty where the stereo matching cost alone fails to capture. These results highlight the advantages of utilizing global multi-view information within MVS. **Figure 10** shows the visualization of the error covariance matrix for each 3D point, propagated using the uncertainty quantification framework integrated with our proposed method. The error covariance matrix represents a transformation of scale and rotation applied to a standard sphere centered at each 3D point. In areas with large uncertainty, e.g., tree regions, cars, and edges of buildings, the ellipsoids are larger than those on ground surfaces or building roofs. The dominant direction of each ellipsoid aligns with the camera viewing direction, indicating that uncertainty is greatest along the Z-axis of the camera coordinate system due to the triangulation process.

### 4.6. Evaluation of Absolute Uncertainty of 3D point clouds

Based on the sensor types, we use typical absolute positioning accuracies: 0.02 m for the UseGeo (UAV platform) dataset and 0.1 m for the Vaihingen (airborne platform) dataset. The results are summarized in **Table 4**. For UseGeo, the performance is similar to the evaluation of relative uncertainty in terms of accuracy, with a higher bounding rate. For Vaihingen, accuracy is lower compared to the relative uncertainty evaluation, although the bounding rate is slightly improved. This is due to the higher altitude, which amplifies errors and makes absolute uncertainty estimation more sensitive to sensor positioning errors. Overall, the experiments demonstrate that our framework can effectively estimate absolute uncertainty, particularly in UAV scenarios. The performance could be further improved if precise absolute positioning accuracies are available from the original device providers.

### 4.7. Sensitivity Analysis

We first analyze the distribution of n-view points across different datasets. As shown in **Figure 11**, although the distributions vary among datasets, all datasets contain a sufficient number of n-view points ($n \geq 6$): 55.40% for UseGeo, 22.83% for Vaihingen, and 39.51% for Dortmund. This is largely due to the overlap ratio in aerial and UAV datasets, which provides sufficient stable points for regressing disparity uncertainty.

Second, we assess the validity of n-view points in our proposed method by evaluating the accuracy of 3D points with respect to the number of views on the UseGeo dataset. As shown in **Figure 12(a)**, the accuracy of multi-view points ($n \geqslant 3$) is significantly better than that of 2-view points, and the accuracy curve flattens after 6 views. This indicates that multi-view points are reliable as pseudo-check points for disparity uncertainty regression. Therefore, while we use $n \geqslant 6$ in our experiments, it remains a tunable hyperparameter and can be reduced to as low as n=3 or 4 for datasets with fewer overlapping images, while still ensuring the validity of our proposed method.

Last, we analyze how different choices of n affect the performance of the method. We use the UseGeo datasets for this sensitivity analysis and evaluate the bounding rates and mean error of our method using n-view points, with n ranging from 4 to 8. As shown in **Figure 12(b)**, as n increases, accuracy improves (i.e., mean error

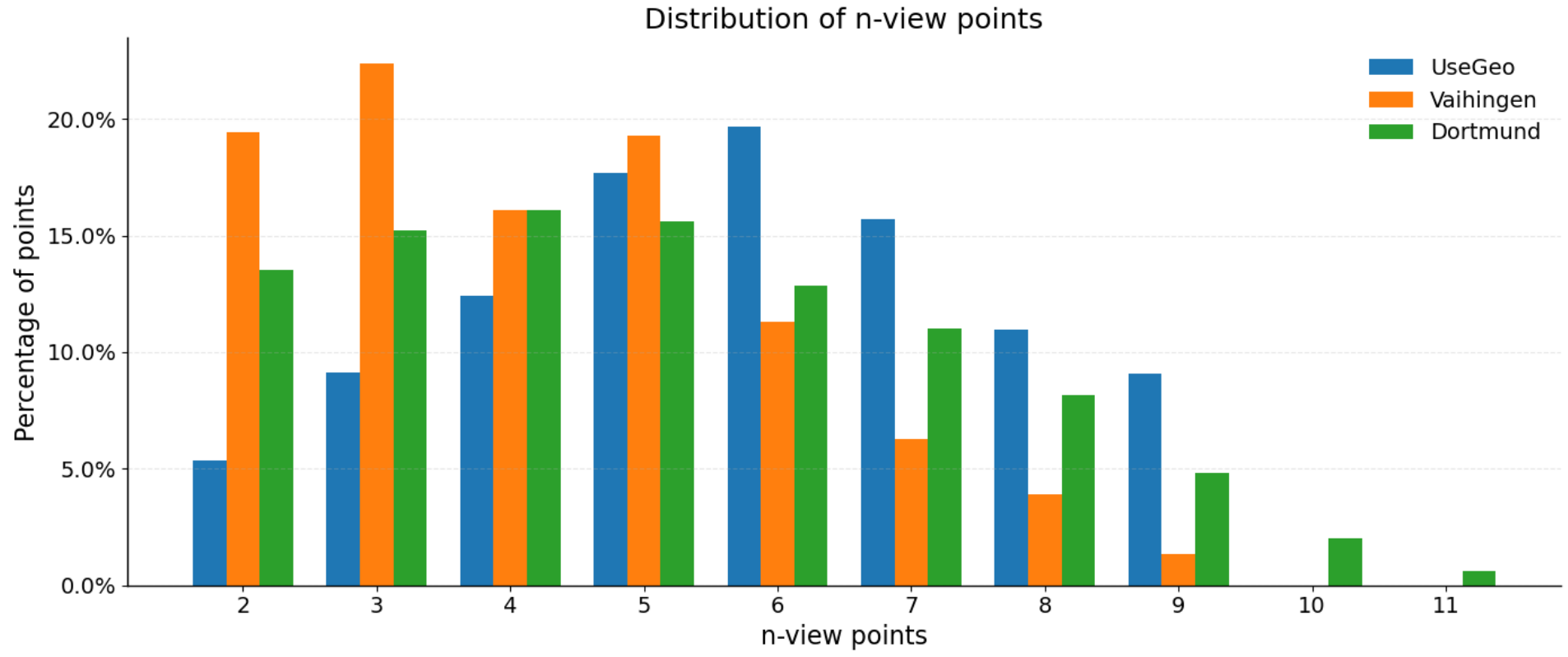

**Figure 11**

Distribution of n-view points across different datasets.

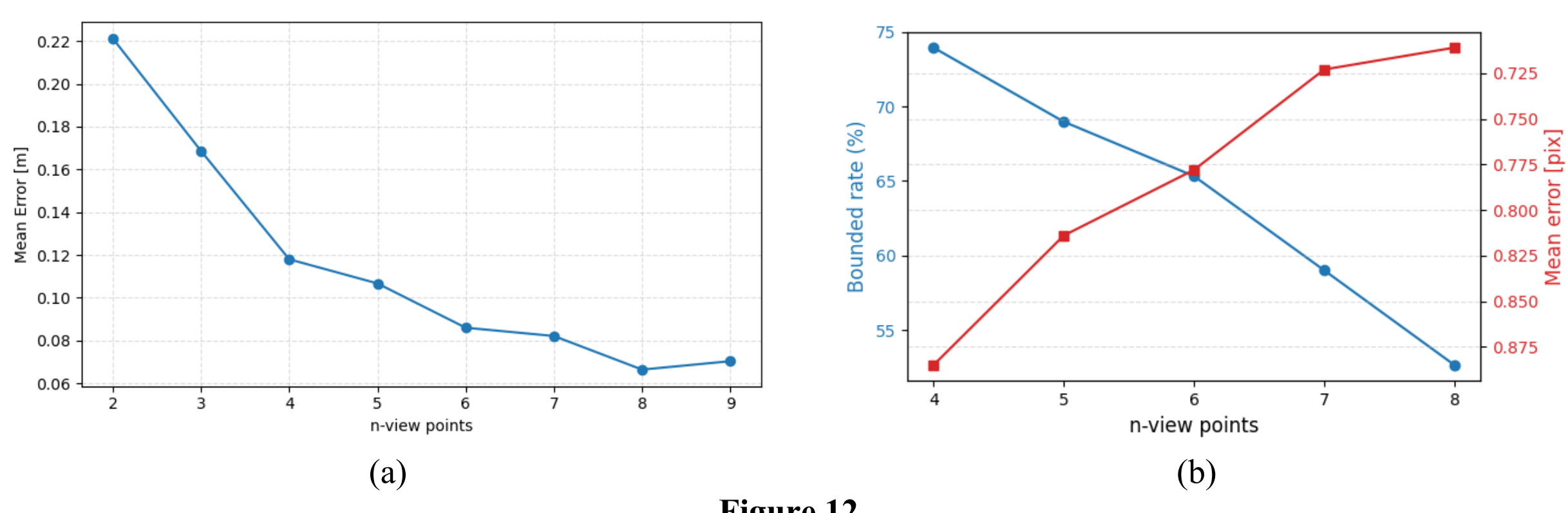

(a) (b)

**Figure 12**

(a) Evaluation of 3D point accuracy with respect to number of views. (b) Sensitivity analysis of the choice of n for n-view points. The vertical axis for mean error is inverted so that lower values appear higher, reflecting the fact that smaller mean errors indicate better performance.

decreases), while the bounding rate decreases. The n-view points (n≥8) achieve the best accuracy (i.e., lowest mean error) due to the use of more accurate points for self-calibrating; however, the number of such points is too small, which leads to an underestimation of uncertainty for most pixels and consequently a poor bounding rate. On the contrary, n-view points (n≥4) provide a more general representation, resulting in the highest bounding rate, but the accuracy degrades due to the inclusion of less accurate points for self-calibrating. The n-view points (n≥6) achieve the best trade-off between bounding rate and mean error. These points are sufficient and reliable, making n≥6 the optimal choice for most scenarios.

## 5. CONCLUSION

This paper presents an uncertainty quantification framework tailored for aerial and UAV photogrammetry. The framework associates each 3D point with an error covariance matrix, which is propagated through the two-step photogrammetry process. A key contribution is a novel, self-calibrating method for estimating uncertainty in the MVS stage. Our method leverages self-contained and stably accurate n-view points (n≥6) from MVS to regress disparity uncertainty using a highly relevant cue (i.e., matching cost) for each stereo pair. Compared to existing approaches, our method is self-supervised and does not require additional training data from external sources, thereby avoiding common pitfalls such as domain shifts and limited transferability. As a result, it achieves improved generalization across diverse scenes. To the best of our knowledge, it is one of the first self-supervised methods that enables statistically grounded error propagation adhering to the photogrammetry process. The performance of the framework is validated on various airborne and UAV datasets. Comprehensive evaluations are conducted both on disparity uncertainty and the uncertainty of 3D point clouds. Results demonstrate the superiority of our method in terms of accuracy and bounding rates, providing uncertainty estimates that closely match actual errors without overestimation. This uncertainty quantification framework delivers robust and certifiable per-point accuracy estimates, facilitating broader adoption of photogrammetric point clouds in downstream applications. One potential limitation of our method lies in the reliance on n-view points ($n \geq 6$) for uncertainty estimation in MVS. While this is generally sufficient in aerial and UAV photogrammetry, it may require re-tuning for datasets with significantly lower image overlap or different acquisition geometries. In such cases, the number of stable n-view points may decrease, potentially affecting the overall performance of the proposed method. Future work could investigate alternative strategies for selecting stably accurate 3D points. Another promising direction is to extend the framework to 3D mesh uncertainty quantification. For evaluation, future research could assess the full covariance matrix by constructing a dataset with known pointwise correspondences between photogrammetric reconstructions and ground truth point clouds.

## ACKNOWLEDGMENT

This work was partially supported by the Office of Naval Research (ONR, Award No. N00014-20-1-2141 & N00014-2312670). The authors would like to acknowledge the provision of the UseGeo datasets by ISPRS Scientific Initiative USEGEO - https://usegeo.fbk.eu. The Vaihingen data set was provided by the German Society for Photogrammetry, Remote Sensing and Geoinformation (DGPF): https://ifpwww.ifp.uni-stuttgart.de/dgpf/DKEP-Allg.html. For the Dortmund dataset, the authors would like to acknowledge the provision of the datasets by ISPRS and EuroSDR, released in conjunction with the ISPRS scientific initiative 2014 and 2015, led by ISPRS ICWG I/Vb.

# Appendix 1

In this appendix, we provide a detailed derivation of the error propagation formulas.

The condition equations for multi-ray intersection or bundle adjustment are shown as follows:

$$F_i = \pi(X, \theta_i) - x_i = \begin{bmatrix}0\\0\end{bmatrix}, \qquad i = 1, \dots, n$$

For this example, $X = [X, Y, Z]^T \in \mathbb{R}^3$ is a 3D point observed in n images, $\theta_i = \left[\theta_{i1}, \theta_{i2}, \dots, \theta_{ip}\right]^T$ denotes the i-th camera parameters (including intrinsics and extrinsics), $x_i = [u_i, v_i]^T \in \mathbb{R}^2$ is the 2D observation in image i, $\pi$ is the projection function, and $F_i$ is the condition function.

These equations can be used in a least-square solution to solve for X. We linearize the condition equations using a first-order Taylor expansion:

$$Av + B_X\Delta = f$$

where A is the $2n \times np$ Jacobian matrix of F with respective to the camera parameters, $B_X$ is the $2n \times 3$ Jacobian matrix of F with respective to 3D point coordinates, f is the $2n \times 1$ misclosure vector, v is the $np \times 1$ residual vector for camera parameters, and $\Delta$ is the $3 \times 1$ correction vector for the 3D point coordinates.

A takes the form shown as follows:

$$\underset{2n \times np}{A} = \begin{bmatrix} A_1 & 0 & \cdots & 0 \\ 0 & A_2 & \cdots & 0 \\ \vdots & \vdots & \ddots & \vdots \\ 0 & 0 & \cdots & A_n \end{bmatrix},$$

$$\underset{2 \times p}{A_i} = \begin{bmatrix} \frac{\partial u_i}{\partial_{\theta_{i1}}} & \frac{\partial u_i}{\partial_{\theta_{i2}}} & \cdots & \frac{\partial u_i}{\partial_{\theta_{ip}}} \\ \frac{\partial v_i}{\partial_{\theta_{i1}}} & \frac{\partial v_i}{\partial_{\theta_{i2}}} & \cdots & \frac{\partial v_i}{\partial_{\theta_{ip}}} \end{bmatrix}$$

where the 0 symbols correspond to the $2 \times p$ zero matrices. It should be noted that, the dimension and structure of A are slightly different if the images share the same camera model, but it can be composed in a similar way.

Similarly, $B_X$ takes the form shown as follows:

$$\underset{2n \times 3}{B_X} = \begin{bmatrix} B_{X1} \\ B_{X2} \\ \vdots \\ B_{Xn} \end{bmatrix}, \qquad \underset{2 \times 3}{B_{Xi}} = \begin{bmatrix} \frac{\partial u_i}{\partial_X} & \frac{\partial u_i}{\partial_Y} & \frac{\partial u_i}{\partial_Z} \\ \frac{\partial v_i}{\partial_X} & \frac{\partial v_i}{\partial_Y} & \frac{\partial v_i}{\partial_Z} \end{bmatrix}$$

In order to solve the linearized condition equations, the set of normal equations are used as follows:

$$B_X^T\left(A\Sigma_\theta A^T + \Sigma_{disp}\right)^{-1} B_X\Delta = B_X^T\left(A\Sigma_\theta A^T + \Sigma_{disp}\right)^{-1} f$$

where $\Sigma_\theta$ is the $np \times np$ covariance matrix of the camera parameters.

$$\underset{np \times np}{\Sigma_\theta} = \begin{bmatrix} \underset{p \times p}{\Sigma_{\theta 1,1}} & \cdots & \underset{p \times p}{\Sigma_{\theta 1,n}} \\ \vdots & \ddots & \vdots \\ \text{sym.} & \cdots & \underset{p \times p}{\Sigma_{\theta n,n}} \end{bmatrix}$$

It is derived using USFM framework [65]. The gauge freedom is fixed by combining constrained Gauss-Markov model with novel approach for nullspace computation. The readers are encouraged to refer to the Polic, et al. [65]'s work for further details.

$\Sigma_{disp}$ is the $2n \times 2n$ covariance matrix constructed from the disparity uncertainties of the n matched pixels:

$$\underset{2n \times 2n}{\Sigma_{disp}} = \begin{bmatrix} u_{x_1}^2 & u_{x_1 y_1} & 0 & 0 & \cdots & 0 & 0 \\ u_{y_1 x_1} & u_{y_1}^2 & 0 & 0 & \cdots & 0 & 0 \\ 0 & 0 & u_{x_2}^2 & u_{x_2 y_2} & \cdots & 0 & 0 \\ 0 & 0 & u_{y_2 x_2} & u_{y_2}^2 & \cdots & 0 & 0 \\ \vdots & \vdots & \vdots & \vdots & \ddots & \vdots & \vdots \\ 0 & 0 & 0 & 0 & \cdots & u_{x_n}^2 & u_{x_n y_n} \\ 0 & 0 & 0 & 0 & \cdots & u_{y_n x_n} & u_{y_n}^2 \end{bmatrix}$$

The estimated disparity uncertainties u from our proposed method, which are initially defined in epipolar space, are projected to the image plane to obtain their x and y components.

The solution is obtained by applying the correction vector $\Delta$ to the initial 3D point estimates:

$$X = X_0 + \Delta$$

The covariance matrix of the 3D point is obtained as follows:

$$\underset{3\times 3}{\Sigma_g} = \begin{bmatrix} \sigma_X^2 & \sigma_{XY} & \sigma_{XZ} \\ & \sigma_Y^2 & \sigma_{YZ} \\ \text{sym.} & & \sigma_Z^2 \end{bmatrix}$$

$$= \left(\Sigma_\varepsilon^{-1} + B_X^T\left(A\Sigma_\theta A^T + \Sigma_{disp}\right)^{-1} B_X\right)^{-1}$$

Where $\Sigma_\varepsilon^{-1}$ is a diagonal matrix with small positive values ($10^{-6}$) to stabilize the solution.